\documentclass[runningheads]{llncs}
\usepackage{graphicx}
\usepackage{amsfonts}
\usepackage{multirow}
\begin{document}
\title{A Sketch-Based System for Semantic Parsing}
%
%
%
\author{Zechang Li\inst{1,2} \and
	Yuxuan Lai\inst{1} \and
	Yuxi Xie\inst{1} \and
	Yansong Feng\inst{1*} \and
	Dongyan Zhao\inst{1,2}
	}
\authorrunning{Z. Li et al.}
%
\institute{Wangxuan Institute of Computer Technology, Peking University, Beijing, China \and
	Center for Data Science, Peking University, Beijing, China \\
	\email{\{zcli18, erutan, xieyuxi, fengyansong, zhaody\}@pku.edu.cn}}
\maketitle              
\begin{abstract}
This paper presents our semantic parsing system for the evaluation task of open domain semantic parsing in NLPCC 2019. Many previous works formulate semantic parsing as a sequence-to-sequence(seq2seq) problem. Instead, we treat the task as a sketch-based problem in a coarse-to-fine(coarse2fine) fashion. The sketch is a high-level structure of the logical form exclusive of low-level details such as entities and predicates. In this way, we are able to optimize each part individually. Specifically, we decompose the process into three stages: the sketch classification determines the high-level structure while the entity labeling and the matching network fill in missing details. Moreover, we adopt the seq2seq method to evaluate logical form candidates from an overall perspective. The co-occurrence relationship between predicates and entities contribute to the reranking as well. Our submitted system achieves the exactly matching accuracy of 82.53\% on full test set and 47.83\% on hard test subset, which is the 3rd place in NLPCC 2019 Shared Task 2. After optimizations for parameters, network structure and sampling, the accuracy reaches 84.47\% on full test set and 63.08\% on hard test subset.\footnote{Our code and data are available at https://github.com/zechagl/NLPCC2019-Semantic-Parsing.}

\end{abstract}
\section{Introduction}

Open domain semantic parsing aims to map natural language utterances to structured meaning representations. Recently, seq2seq based approaches have achieved promising performance by structure-aware networks, such as sequence-to-action\cite{chen2018sequence} and STAMP\cite{sun-etal-2018-semantic}.

However, this kind of approach mixes up low-level entities, predicates and high-level structures together, which loses precision at each level to some extent. So the sketch-based method may be an another choice for disentangling high-level structures from low-level details. In this work, we conduct our sketch-based approach on MSParS, a large hand-annotated semantic dataset mapping questions to logical forms. We argue there are at least two advantages to sketch-based method. Firstly, basic attention based seq2seq network\cite{bahdanau2014neural,luong-etal-2015-effective} does not perform well in semantic parsing because logical forms are structured sequences and it fails to incorporate structure information of logical forms. Then sequence-to-tree(seq2tree)\cite{dong-lapata-2016-language} proposes a structure-aware decoder to utilize the information. But its architecture also becomes much more complex. Instead of using intricate decoders, we can extract high-level sketches for logical forms and classify samples into several sketch classes. Logical forms of a certain sketch class have a fixed pattern which is shown in Table~\ref{table1}. So the structure problem is finally simplified to a classification task. Secondly, logical forms often need to copy a span of questions. Although Copynet\cite{gu-etal-2016-incorporating} and Pointer\cite{see-etal-2017-get} implement the copy mechanism, it is still difficult to achieve the expected effect. But for the sketch-based method, this problem becomes an individual entity labeling task which is easier than generating entities. Generally speaking, the seq2seq way decodes the entire meaning representation in one go while we deal with different parts at different levels of granularity just like coarse2fine\cite{dong-lapata-2018-coarse}. Although we increase the number of stages, the network architecture of each stage is much simpler without sacrificing the accuracy. In this way, we are able to locate the errors and optimize according parts.

\begin{table}
	\caption{Examples demonstrating sketches of logical forms. $P$ represents predicate and $E$ represents entity. Subscripts are applied to distinguish different ones.}\label{table1}
	\centering
	\begin{tabular}{l|p{250pt}}
		\hline
		Class           & Sketch of Logical Form \\ \hline
		aggregation     & count ( lambda ?x ( $P_1\;E_1$ ?x ) ) \\ \hline
		cvt & ( lambda ?x exist ?y ( and ( $P_1\;E_1$ ?y ) ( $P_2$ ?y $E_2$ ) ( $P_3$ ?y ?x ) ) ) \\ \hline
		multi-turn-entity & ( lambda ?x ( $P_1\;E_1$ ?x ) ) $|||$ ( lambda ?x ( $P_2\;E_1$ ?x ) ) \\ \hline
		multi-turn-answer$\ $ & ( lambda ?x ( $P_1\;E_1$ ?x ) ) $|||$ ( lambda ?x exist ?y ( and ( $P_1\;E_1$ ?y ) ( $P_2$ ?y ?x ) ) ) \\ \hline
		single-relation & ( lambda ?x ( $P_1\;E_1$ ?x ) ) \\ \hline
		yesno       & ( $P_1\;E_1\;E_2$ ) \\ \hline
	\end{tabular}
\end{table}

We propose to decompose the process into three stages. In the first stage, we deal with a sketch classification task. Then, we find the entities in the questions through an entity labeling task. Actually, we combine the two stages through the multi-task model for both accuracy and efficiency\cite{chen2019bert}. The last stage is the most difficult part since the knowledge base of MSParS is not available. We define question pattern-logical form pattern pair and use the matching network to rank all these pairs. Seq2seq based approach is one of the two methods we adopted here to help rescore on the whole. We also incorporate state-of-art pre-trained work, Bert\cite{devlin-etal-2019-bert}, in above tasks to incorporate more priori knowledge.

The error rate of our multi-task model is lower than 2\%, which ensures the right sketch and entities. So the last stage actually determines the accuracy to a large extent. Our accuracy achieves 77.42\% after above three stages. Seq2seq based approach and co-occurrence relationship improve the accuracy to 86.86\% in validation set. Our final accuracy in full test set reaches 84.47\%. And the accuracy on hard test subset has been promoted to 63.08\% finally which is higher than the best model on the submission list by 5.65\%.

In the rest of our paper, we first analyze the special features of MSParS for this task in section 2. Afterwords, we discuss our system in detail in section 3. Then in section 4, we demonstrate our experimental setup, results and analyzation. Related works are mentioned in section 5. At last, we make a conclusion of the whole paper and propose our future work.

\section{Data Analyzation}

The dataset MSParS is published by NLPCC 2019 evaluation task. The whole dataset consists of 81,826 samples annotated by native English speakers. 80\% of them are used as training set. 10\% of them are used as validation set while the rest is used as test set. 3000 hard samples are selected from the test set. Metric for this dataset is the exactly matching accuracy on both full test set and hard test subset. Each sample is composed of the question, the logical form, the parameters(entity/value/type) and question type as the Table~\ref{table5} demonstrates.

\begin{table}
	\caption{An sample of MSParS.}\label{table5}
	\centering
	\begin{tabular}{l|l}
		\hline
		question           & what is birth date for chris pine \\ \hline
		logical form     & ( lambda ?x ( mso:people.person.date\_of\_birth chris\_pine ?x ) ) \\ \hline
		parameters & chris\_pine (entity) [5,6] \\ \hline
		question type$\ $ & single-relation \\ \hline
	\end{tabular}
\end{table}

Samples are classified to 12 classes originally at a coarse level while we reclassify them at a finer level, which is the basis of our sketch-based method. We replace the predicate in the triple as $P_i$, the entity in the triple as $E_i$ and distinguish different ones with subscripts. The number in superlative class and comparative class is replaced as $V$ while the type in the triple begin with special predicate ``isa" is replaced as $T$ as well. In this way, we get the sketch of the logical form. Finally, we produce 15 classes of sketches.

We believe the features of questions highly correlate with the sketch of logical forms. For instance, the sketch must begin with ``argmore" or ``argless" if there are comparative words such as ``higher", ``more" and ``before" in questions. Therefore, we take questions as input to classify samples to different sketch classes.

As the Table~\ref{table5} suggests, entities are concatenated tokens from the question. So we implement entity labeling to label every token in the questions.

Nonetheless, cases are tough when there are more than one entities in the logical form. Suppose that we have labeled $E_1$ and $E_2$ from the question. We do not know which one we should choose to fill in the first entity slot in the sketch. We solve this problem and pick out the suitable predicate simultaneously. The entities in the questions are replaced by label ``entity'' with subscipts suggesting the order they appear in questions to get question patterns. When it comes to logical form patterns, the entities in logical forms are substituted as well while predicates are split to small tokens. Table~\ref{table2} gives an example of these two patterns. In this way, we combine the entity collocations with predicates successfully. Another reason for label ``entity'' used here is generalization. For instance, ``what is birth date for barack obama" shares the same question pattern ``what is birth date for entity1" with ``what is birth date for donald trump". The predicate used in these logical forms is ``mso:people.person.date\_of\_birth''. So we can draw the conclusion that the predicate for this question pattern is likely to be ``mso:people.person.date\_of\_birth''. If ``what is birth date for george bush" appears in the test set, we are able to find the right predicate even if we do not see ``george bush'' before. Without the impact of specific entities, our model learns the mapping from question patterns to logical form patterns more accurately. Since we do not have a knowledge base, we can only extract logical form patterns in training set. And we find 90.34\% of logical form patterns in validation set are covered by that in training set, which ensures the feasibility of our method.

\begin{table}
	\caption{An example for question pattern and logical form pattern.}\label{table2}
	\centering
	\begin{tabular}{l|p{220pt}}
		\hline
		question & travels in the interior districts of africa has how many pages? $|||$ when is the date of publication of the book edition? \\ \hline
		question pattern & entity1 has how many pages? $|||$ when is the date of publication of the book edition? \\ \hline
		logical form & ( lambda ?x ( mso:book.edition.number\_of\_pages travels\_in\_the\_interior\_districts\_of\_africa ?x ) ) $|||$ ( lambda ?x ( mso:book.edition.publication\_date travels\_in\_the\_interior\_districts\_of\_africa ?x ) ) \\ \hline
		logical form pattern$\ $ & book edition number of pages entity1 ?x $|||$ book edition publication date entity1 ?x \\ \hline
	\end{tabular}
\end{table}

We take question patterns paired with logical form patterns as input. Then, we get logical form candidates through combining sketches and entities with logical form patterns. The ones with higher scores are more likely to be right.

\section{Proposed Approach}
\subsection{Sketch Classification}

The single sentence classification fine-tuned task in Bert is applied in this stage. A special classification embedding ([CLS]) is added to the beginning. We use the final hidden state corresponding to this token as the aggregate sequence representation for classification task denoted as $C_s \in \mathbb{R}^h$, so the probability of class $c_i$ can be computed as:
\begin{equation}
p(c_i|x) = softmax_i(W_sC_s + b_s)
\end{equation}
where $W_s \in \mathbb{R}^{k_s \times h}$ and $b_s \in \mathbb{R}^{k_s}$, $k_s$ is the number of sketch classes here. $W_s$, $b_s$ and all the parameters of Bert are fine-tuned jointly to maximize the log likelihood probability of the correct label.

\subsection{Entity Labeling}

We use the single sentence tagging fine-tuned task in Bert here to label every token in the question whether it is an entity token that appears in the logical form as well. To simplify the problem, we use 3 labels for the tokens in the questions. Label ``b" represents the first token in an entity while label ``i'' for the rest ones. And label ``o'' represents those tokens which are not in any entities. Because of the lexical rules in Bert, we also label the special token ([CLS]) at the beginning of the sentence and the special token ([SEP]) at the ending of the sentence as ``o''. The last label ``p'' is for all the padding tokens added to reach max\_length. Besides, some tokens in the questions are split into several smaller tokens by Bert. For the split ones, they are labeled as ``i'' if they are in the entities and ``o'' otherwise. In this stage, we use all the final hidden states denoted as $D \in \mathbb{R}^{h \times m}$ where m is the max\_length of the input tokens we set. The hidden state is mapped into dimension $k_e$ via $E = W_eD + b_e$ where $W_e \in \mathbb{R}^{k_e \times h}$ and $b_e \in \mathbb{R}^{k_e \times m}$, $k_e$ is the number of labels here. We employ the CRF on the top of the network taking $E$ as input representations. The objective is to minimize the loss of CRF layer.

\subsection{Multi-Task Model}

\begin{figure}
	\includegraphics[width=\textwidth]{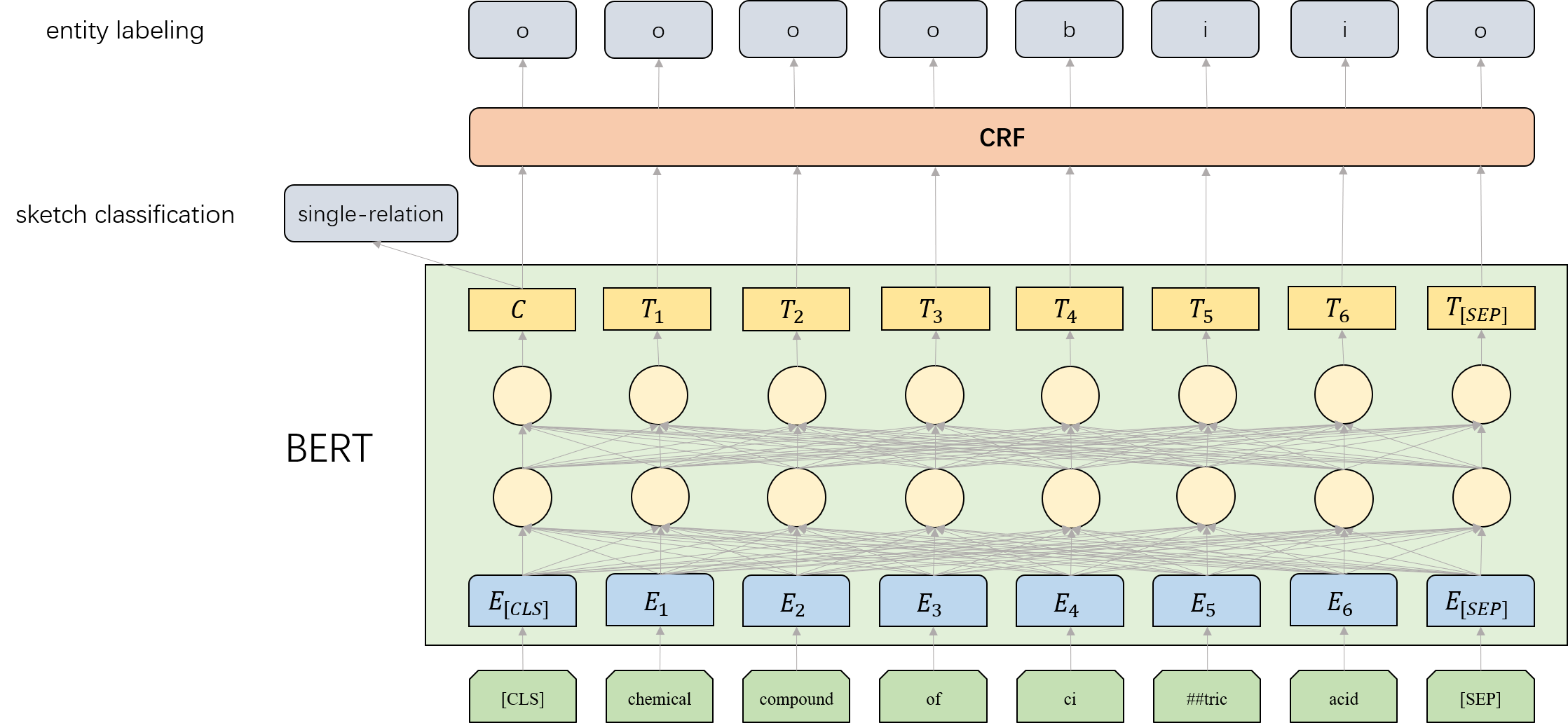}
	\caption{An overview of multi-task model proposed. The original input question is ``chemical compound of citric acid". It becomes ``chemical compound of ci \#\#tric acid" after the tokenization of Bert.} \label{fig1}
\end{figure}

We combine sketch classification and entity labeling to share information together, which means sketches of samples can help label entities while the labeled entities can help sketch classification conversely. The architecture of our model is shown in Fig.~\ref{fig1} where the parameters of Bert model is fine-tuned together for two tasks. Since the scale of dataset is large, we can save lots of time through multi-task model instead of training two different models. Finally, it contributes to both accuracy and efficiency. In this way, our loss to minimize is the weighted sum of the cross-entropy loss in sketch classification task and the CRF loss in entity labeling task.

\subsection{Pattern Pair Matching Network}

Besides the single sentence tasks, Bert provides sentence pair classification tasks as well. We implement the matching network taking question patterns and logical form patterns as input. The right pattern pairs are regarded as positive samples. We select negative samples only from the logical form patterns in the same sketch class for fixed question patterns. The sketch mentioned is from the multi-task model. Just like sketch classification, we denote the final hidden state corresponding to token ([CLS]) as $C_p \in \mathbb{R}^h$, so the probability can be computed as:
\begin{equation}
p(c_j|x) = softmax_j(W_pC_p + b_p)
\end{equation}
where $W_p \in \mathbb{R}^{2 \times h}$, $b_p \in \mathbb{R}^{2}$ and $c_j \in \{0, 1\}$. $W_p$, $b_p$ and all the parameters of bert are fine-tuned jointly to maximize the log likelihood probability of the correct class.

In the prediction stage, the candidates for a question pattern are from logical form patterns in the same sketch class as well. The probabilities of class ``1" are scores we get for these pattern pairs. From logical form patterns, we get not only right predicates, but right orders as well in which entities should appear. So with the sketch and entities we aquire in the multi-task model, we can already generate complete logical form candidates with scores between 0 and 1.

\subsection{Predicate-Entity Pair Matching Network}

To alleviate the absence of knowledge base, we incorporate the co-occurrence relationship between predicates and entities to evaluate the candidates. We create the second matching network based on Bert as well. This time, the pairs we take as input are predicate-entity ones. We label the predicate-entity pair as ``1'' if they have ever appeared in one triple in training set. For a certain entity, we select predicates that never appear with this entity as negetive samples. In the prediction stage, we score the predicate-entity pairs in logical form candidates. However, this network does not take questions into account. The predicate for a certain entity can differ a lot according to various questions. For instance, the predicate for ``what is birth date for barack obama" is apparently different from that for ``what is birth place for barack obama". But the entity ``barack obama" has only one predicate with highest score. Although this matching network only considers the co-occurrence relationship regardless of the information from questions, scores produced by it do work as an auxiliary.

\subsection{Pointer Network}

Although it is not easy for a complex network to generate the whole logical form, such networks do reflect the mapping from an overall perspective. So we adopt Pointer\cite{see-etal-2017-get} here to rerank. We take the questions as input. For logical forms, entities and predicates are composed of words concatenated by ``\_'' or ``.''. In order to utilize the information of words, we split all entities and predicates and take split logical form candidates as output. For a fixed question, we calculate cross-entropy losses of different pairs with split logical forms. Then every loss is divided by the max one and subtracted by 1 to be normalized between 0 and 1. The higher the score is, the more the logical form candidate is likely to be true.

\subsection{Ranking}

A linear combination of the three intermediate scores from pattern pair matching network, predicate-entity pair matching network and Pointer is used to rerank logical form candidates. Weights are roughly adjusted in validation set.

\section{Experiment}

The number of sketch classes is 15 and the number of labels is 4 in the multi-task model. The Bert model we applied is ``BERT-Base, Uncased" with 12-layer, 768-hidden, 12-heads and 110M parameters\footnote{https://github.com/google-research/bert}. All the parameters are fine-tuned in validation set. In the multi-task model, we train the model for 10 epoches. We set batch size to 32 and learning rate to 2e-5. The weight of the loss in sketch classification is 1 while that in entity labeling is 2. We train 3 models in pattern pair matching network with different epoches. As for predicate-entity pair matching network, the number of epoch we use is 3. In Pointer, word embeddings were initialized by Glove\cite{pennington2014glove}. The hidden dim of LSTM is set to 256. More details will be released in our source codes later.

Because of the instability of the performance of neural network over training epoches,
ensemble learning is incorporated both in pattern pair matching network and Pointer. Scores of Pointer is the simple average of scores from 3 models with different epoches. When it comes to pattern pair matching net, it is a little complex. We make a prediction for training set with our ``best" model. We apply ranking sampling here. From those labeled as ``0'' but with probabilities larger than 0.0001, we select 20 of them while 5 of those whose probabilities are smaller than 0.0001 as new negative samples. We train new models with new training data resampled before every epoch based on one ``best" model and base model of Bert. After several epoches, we average the probabilities of new models and original models for ensemble.

We demonstrate the detailed performance in Table~\ref{table3}. All samples are classified into 15 classes. We show the results for every class and the overall weighted average preformance in validation set. Because the complete test set is not open yet, we only provide the overall results in test set returned after submission.

It can be seen the overall error rate of our multi-task model is only 1.93\% which means this task is successful. In sketch classification, $Err_s$ scores of all classes are lower than 1\% except multi-turn-answer. Its recall is 100.00\% while its precision is 91.38\%. 0.92\% of samples in multi-turn-entity are misclassified to multi-turn-answer in validation set. We find there are separator ``$|||$'' in logical forms from three classes of multi-turn questions. Multi-turn-predicate questions have two different entities while both multi-turn-entity and multi-turn-answer questions have only one. This kind of entity information is passed to sketch classification through shared parameters. So our system makes some mistakes while distinguishing multi-turn-entity samples from multi-turn-answer samples. As for entity labeling, the overall error rate is 1.72\%. We check the wrong samples and find our model is not so good at recognizing entity boundaries especially while encountering some special tokens such as articles, genitive ``s'' and quotation mark. Actually, it is not easy for human to define an entity in these cases as well.

\begin{table}
	\caption{Performances of our best model. $F1_s$ represents the $F1$ score of sketch classification in multi-task model. We compute $Err_s$ by $1 - F1_s$. $Err_e$ represents the error rate of entity labeling part in multi-task model(an sample is regarded as right only when all of its entities are labeled correctly). $Err_m$ represents the error rate of the whole multi-task model(an sample is regarded as right only when both sketch classification subtask and entity labeling subtask are correct). $Acc_l$ is the exactly matching accuracy of logical forms. We compute $Err_l$ by $1 - Acc_l$.}\label{table3}
	\centering
	\begin{tabular}{l|l|cccc}
		\hline
		\multicolumn{2}{l|}{Dataset}                 & $\ Err_s\ $   & $\ Err_e\ $ & $\ Err_m\ $ & $\ Err_l\ $ \\ \hline
		\multirow{16}{*}{Dev} & aggregation          & 0.22\%  & 1.99\%  & 2.10\%  &  23.84\%  \\
		& comparative          & 0.00\% & 0.00\%  & 0.00\%  & 0.00\% \\
		& cvt                  & 0.57\%  & 4.04\%  & 4.04\%  & 13.41\% \\
		& multi-choice         & 0.37\%  & 6.72\%  & 7.46\%  & 52.24\% \\
		& multi-constraint     & 0.17\%  & 1.71\%  & 1.71\%  & 6.83\% \\
		& multi-hop            & 0.26\%  & 1.28\%  & 1.54\%  & 3.85\% \\
		& multi-turn-answer    & 4.50\%  & 0.00\%  & 0.00\%  & 5.66\% \\
		& multi-turn-entity    & 0.51\%  & 1.37\%  & 2.29\%  & 16.59\% \\
		& multi-turn-predicate$\ \ $ & 0.50\%  & 2.00\%  & 2.00\%  & 12.00\% \\
		& single-relation      & 0.27\%  & 1.18\%  & 1.31\%  & 9.26\% \\
		& superlative0         & 0.40\%  & 3.04\%  & 3.04\%  & 24.21\% \\
		& superlative1         & 0.00\% & 0.00\%  & 0.00\%  & 6.90\% \\
		& superlative2         & 0.00\% & 0.00\%  & 0.00\%  & 0.00\% \\
		& superlative3         & 0.00\% & 0.00\%  & 0.00\%  & 0.00\% \\
		& yesno                & 0.17\%  & 1.33\%  & 1.33\%  & 9.67\% \\ \cline{2-6}
		& overall              & 0.36\%  & 1.72\%  & 1.93\%  & 13.14\% \\ \hline
		\multicolumn{2}{l|}{Test(full)}                    & -        & -       & -       & 15.53\% \\ \hline
		\multicolumn{2}{l|}{Test(hard)}                    & -        & -       & -       & 36.92\% \\ \hline
	\end{tabular}
\end{table}

At last, $Err_f$ of our best model is 13.14\% in validation set, 15.53\% in full test set and 36.92\% in hard test subset. We inspect the output of our model in order to identify the causes of errors. The entity error takes up 20.43\% not only because of wrong entities but also right entities in wrong order. 79.57\% of incorrect samples have wrong predicates although their entities are right. Our accuracy is extremely low for multi-choice. We look into this class and find 50.72\% of errors are because of right entities with wrong order. Actually, there are three different entities in sketch of multi-choice class and two of them are semantically exchangeable in the form $( or ( equal\ ?x\ E_1 ) ( equal\ ?x\ E_2 ) )$. So it is not easy for our pattern pair matching network to deal with this problem. In the meantime, our model achieves error rate of 0\% for 3 classes in validation set.

\begin{table}
	\caption{Performance comparison. Metric is exactly matching accuracy of logical forms. ``Cover" represents our covered words supplementry. ``Point" represents the application of Pointer losses. ``Pep'' represents the predicate-entity pair matching network.}\label{table4}
	\centering
	\begin{tabular}{l|ccc}
		\hline
		\multirow{2}{*}{System}            & \multicolumn{3}{c}{Acc} \\ \cline{2-4} 
		& Dev        & Test(full)  & Test(hard)     \\ \hline
		$Soochow\_SP(1st)$                      & -          & {\bfseries 85.68\%}  & 57.43\%  \\
		$NP$-$Parser(2nd)$                        & -          & 83.73\% & 51.93\%   \\
		$WLIS(Ours)(3rd)$                       & 82.86\%    & 82.53\%  & 47.83\%  \\
		$BBD(4th)$                      & -          & 68.82\% & 35.41\%   \\ \hline
		$WLIS_{NEW}(Our\ New\ Baseline)\ $         &  77.42\%   & -  & -    \\
		$WLIS_{NEW} + point$               &  84.02\%   & -  & -    \\
		$WLIS_{NEW} + pep$                 &  85.99\%   & -  & -    \\
		$WLIS_{NEW} + point + pep\ $ &  {\bfseries 86.86\%}   &  84.47\% & {\bfseries 63.08\%} \\ \hline
	\end{tabular}
\end{table}

Our system is compared with that of other teams in NLPCC 2019 Shared Task 2. The top 4 results are shown in Table~\ref{table4}. Our system on the submission list is $WLIS$ which achieves the 3rd place. After some optimizations for parameters, seq2seq network structure and sampling, the performance of our new system has been improved a lot. The accuracy of our new baseline reaches 77.42\%. By incorporating two auxiliary scores, the accuracy is improved to 86.86\% in validation set. Accuracy achieves 84.47\% in full test set and 63.08\% in hard test subset. Our accuracy in full test set supasses the 2nd place but is still lower than the 1st place by 1.21\% while the accuracy on hard subset is higher than that of the 1st place by 5.65\%.

\section{Related Work}

Semantic parsing is a long-standing problem in NLP mapping natural language utterances to logical forms\cite{berant2014semantic,kwiatkowski2011lexical,liang2013learning,pasupat2016inferring,wong2007learning,zettlemoyer2005learning}. Since it is not easy for semantic parsing to label data manually, reinforcement learning\cite{liang2018memory} and transfer\cite{wang2018multi,xiong2019transferable} are applied when data is not enough. But in most cases, we are studying how to improve the results when enough data is available for supervised learning. Basic seq2seq network\cite{NIPS2014_5346} enables the model to be trained in an end-to-end mode. Later, structure-aware models are designed to generate logical forms more elaborately. Seq2tree\cite{dong-lapata-2016-language} is equipped with a tree-structured decoder to parse hierarchical logical forms while STAMP\cite{sun-etal-2018-semantic} adopts a switching gate in the decoder to control the generation of SQL. The models mentioned above all generate the whole logical form in one go.

There are also some works that applied sketch-based approach to solve the problem. It has already been explored in the field of program synthesis\cite{solar2008program}. Coarse2fine\cite{dong-lapata-2018-coarse} decomposes the decoding process to 2 stages. Sketches are generated in the first stage while model in the second stage fills in missing details. SQL generating is especially suitable for this method because of its easy sketches. Sqlnet\cite{xu2017sqlnet} divides the task into 6 subtasks to generate different part of SQL. SQLova\cite{hwang2019comprehensive} also inherits this idea and incorporate Bert\cite{devlin-etal-2019-bert} in his model. The idea of our system is similar to that of SQLova. We do not use complex decoders to make our network structure-aware. The architectures of models are easy in every stage. We first determine sketches as the high-level structure. Low-level details are added in later stages. The losses of seq2seq network is applied here to rerank from an overall perspective. So we actually combine both seq2seq method and sketch-based method to some extent.

\section{Conclusion}

In this paper, we presented a sketch-based system for semantic parsing which disentangles high-level structures from low-level details. Due to the absence of knowledge base, we propose to collect question patterns and logical form patterns to capture the implicit relationship between questions and predicates, which can then be used to  perform reranking in a Pointer network within a seq2seq framework. Our previous submitted system achieves the 3rd place while our new system outperforms the 1st place for accuracy in hard test subset. Since the knowledge base will be released later, in future work we would like to incorporate new knowledge to improve our system. We will extend our system to other semantic parsing tasks as well.

\section{Acknowledgements}

This work is supported in part by the NSFC (Grant No.61672057, 61672058, 61872294), the National Hi-Tech R\&D Program of China (No. 2018YFB1005100). For any correspondence, please contact Yansong Feng.

\bibliographystyle{splncs04}
\bibliography{mybibliography}

\end{document}